\documentclass[journal]{IEEEtran}

\usepackage{subcaption}
\usepackage{cite}
\usepackage{amsmath,amssymb,amsfonts}
\usepackage{graphicx}
\usepackage{textcomp}
\usepackage{xcolor}

\usepackage{cite}
\usepackage{amsmath,amssymb,amsfonts}
\usepackage{graphicx}
\usepackage{textcomp}
\usepackage{amsmath}
\usepackage{algpseudocode}
\usepackage{xcolor}
\usepackage[linesnumbered, ruled]{algorithm2e}

\begin{document}
\title{On-Demand Model and Client Deployment in Federated Learning with Deep Reinforcement Learning}
\author{Mario Chahoud$^{1,2,3}$\IEEEauthorrefmark{1}, Hani Sami$^{1,3}$\IEEEauthorrefmark{2}, Azzam Mourad$^{4,3}$\IEEEauthorrefmark{3}, Hadi Otrok$^{5}$\IEEEauthorrefmark{4}, Jamal Bentahar$^{4,1}$\IEEEauthorrefmark{5}, and Mohsen Guizani$^{2}$\IEEEauthorrefmark{6} \\
	\normalsize $^{1}$Concordia Institute for Information Systems Engineering, Concordia University, Montreal, Canada  \\
 \normalsize $^{2}$Mohammad Bin Zayed University of Artificial Intelligence, Abu Dhabi, UAE\\
    \normalsize $^{3}$Artificial Intelligence \& Cyber Systems Research Center, Department of CSM, Lebanese American University, Beirut, Lebanon  \\
	\normalsize $^{4}$KU 6G Research Center, Department of CS, Khalifa University, UAE\\
	\normalsize $^{5}$Center of Cyber-Physical Systems (C2PS), Department of CS, Khalifa University, Abu Dhabi, UAE\\

Email: \IEEEauthorrefmark{1}mario.chahoud@concordia.ca,
\IEEEauthorrefmark{2}hani.sami@concordia.ca,
\IEEEauthorrefmark{3}azzam.mourad@lau.edu.lb,
\IEEEauthorrefmark{4}hadi.otrok@ku.ac.ae,\\
\IEEEauthorrefmark{5}bentahar@ciise.concordia.ca, 
\IEEEauthorrefmark{6}mguizani@ieee.org
}
\maketitle

\begin{abstract}

In Federated Learning (FL), the limited accessibility of data from diverse locations and user types poses a significant challenge due to restricted user participation. Expanding client access and diversifying data enhance models by incorporating diverse perspectives, thereby enhancing adaptability. 
However, challenges arise in dynamic and mobile environments where certain devices may become inaccessible as FL clients, impacting data availability and client selection methods. To address this, we propose an On-Demand solution, deploying new clients using Docker Containers on-the-fly. Our On-Demand solution, employing Deep Reinforcement Learning (DRL), targets client availability and selection, while considering data shifts, and container deployment complexities. It employs an autonomous end-to-end solution for handling model deployment and client selection. The DRL strategy uses a Markov Decision Process (MDP) framework, with a Master Learner and a Joiner Learner. 
The designed cost functions represent the complexity of the dynamic client deployment and selection. 
Simulated tests show that our architecture can easily adjust to changes in the environment and respond to On-Demand requests. This underscores its ability to improve client availability, capability, accuracy, and learning efficiency, surpassing heuristic and tabular reinforcement learning solutions.
\end{abstract}


% \fi
\IEEEpeerreviewmaketitle
\section{Introduction}

\begin{figure}[]
	\centering
	\includegraphics[width=0.45\textwidth]{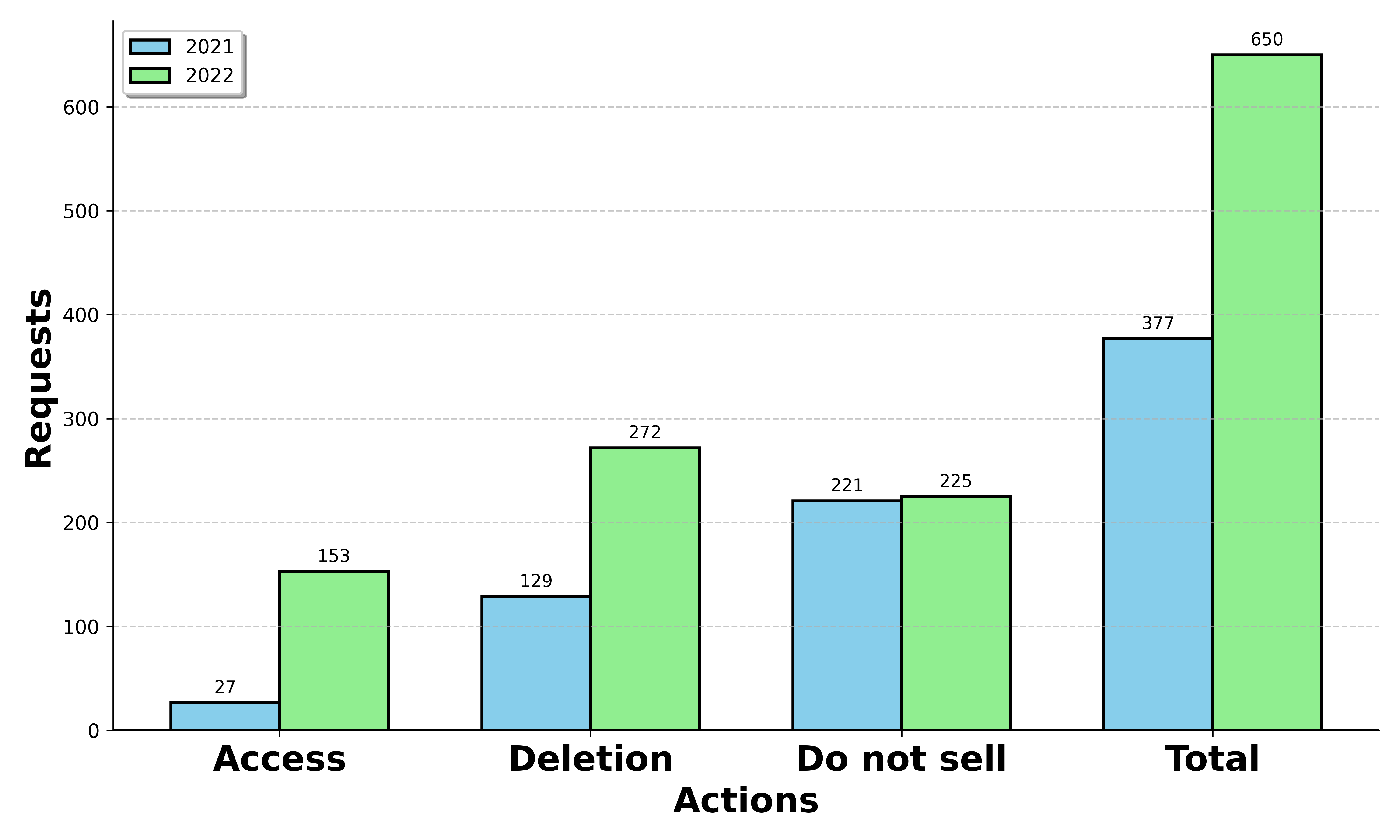}
	\caption{2021 vs. 2022 Volume of Privacy Requests Per Million \cite{WinNT}}
	\label{fig87}
\end{figure}

Various regulations, such as the General Data Protection Regulation in the European Union, aim to protect data privacy \cite{privacyy1}. However, the stringency of these regulations varies globally. A study \cite{WinNT} revealed a notable increase in privacy requests from 2021 to 2022, indicating growing concerns about personal data protection. Access and Deletion requests saw a substantial peak, with a 72\% year-over-year increase in data subject requests (DSRs) per 1 million identities as shown in Figure \ref{fig87}. A common approach to perform learning on such private data, and benefiting from shared knowledge is to use the Federated Learning (FL) mechanism \cite{addkhlm1}. FL is a decentralized machine learning (ML) approach that trains models across local devices, and then aggregates them into a global model, ensuring privacy. Applications such as healthcare monitoring systems, traffic management, and edge computing benefit significantly from FL. %but face substantial challenges due to the dynamic nature of their client environments.
For example, in remote patient monitoring, wearable devices generate health data continuously as patients move. %, requiring rapid decisions in selecting clients for model updates. 
Similarly, in traffic management and autonomous vehicles, FL can enhance traffic prediction models using real-time data from vehicles moving on the road. %, but the fast movement of vehicles demands quick decisions to adapt to changing conditions.
Likewise, edge computing applications rely on FL for collaborative model training across heterogeneous edge devices with high mobility.%, but the mobility of devices and network fluctuations pose challenges in maintaining model consistency.

%Numerous regulations and legal frameworks have been established to safeguard data privacy, exemplified by prominent measures such as the General Data Protection Regulation implemented within the European Union \cite{privacyy1}. Nevertheless, the rigor and strictness of these regulations exhibit notable variations across different nations and geographical regions. Figure \ref{fig87} shows the privacy requests during the periods of 2021 and 2022 \cite{WinNT}. The findings observed a substantial surge in the overall volume of requests compared to the previous year. This underscores the pressing need individuals feel to safeguard their personal information. Experiencing a significant hike in both Access and Deletion requests, the total number of data subject requests (DSRs) per 1 million identities experienced a remarkable 72\% year-over-year increase from 2021 to 2022. Businesses may anticipate receiving approximately 650 requests per 1 million identities, encompassing Access, Deletion, and Do Not Sell requests.

One of the main limitations in existing FL frameworks is in accessing the full potential of available data due to reliance on static clients, leading to incomplete or biased dataset representations and affecting model performance. 
In today's digital landscape, acquiring more clients is about efficiency. Advancements in mobile technology, exemplified by Apple's move towards on-device ML execution \cite{inton1}, signify a transformative leap for resource-constrained devices. This shift underscores devices' capability to independently execute complex ML tasks, reducing reliance on external servers. Docker Containers \cite{inton9} allow swift service deployment on fog/edge devices \cite{M7}, enabling high scalability. They also facilitate distributing ML models directly to clients, explored in our paper as On-Demand FL. Embracing containers for real-time deployment of learning algorithms unlocks new avenues for personalized services and adaptive decision-making.
Docker containers play a crucial role in enabling volunteering devices to participate actively as FL clients. 
Deploying containers that encapsulate all the requisite components for clients, including libraries, methods, and APIs, empowers these devices to seamlessly transition into FL clients. By bundling necessary modules alongside their inherent resources and local data, these containers enable devices to effectively participate in FL processes, contributing to the learning process as clients without the need for extensive setup or configuration. By leveraging Docker containers, resource-constrained devices can dynamically acquire ML capabilities, actively participating in FL \cite{inton2}. This maximizes their computing power utilization and taps into local data without compromising their limited resources. The combination of on-device ML execution, Docker containerization, and FL participation represents a reliable approach for preparing constrained devices for sophisticated ML tasks.
%streamlines the process, facilitating devices to seamlessly transition into FL clients . 
%However, real-world scenarios involve dynamic device availability, varied capabilities, and network conditions, challenging the effectiveness of static deployment strategies \cite{fixx2}.
%Containerization technology allow new devices to function as deployed clients with ML capabilities while resolving issues related to static clients, heterogeneity, and data volume. 

Another major issue is the difficulty in responding effectively to changes in the underlying data distribution \cite{kl4} and adapting to data shifts \cite{kl3} specifically in environments where clients have mobility, impacting model generalization. Researchers propose adaptive learning rate techniques and frameworks, incorporating advanced client selection strategies to optimize data distribution, client selection, and dynamic shifts within the data \cite{datashift1}. Nevertheless, FL's effectiveness relies on having a sufficient number of available clients and not only counting on static clients. Therefore, implementing an On-Demand client deployment technique ensures the availability of new clients for learning and introduces an innovative approach to selecting and managing data shifts. This autonomous, intelligent, and adaptive method transforms new client deployment, providing a dynamic solution that enhances adaptability and responsiveness to changing data distributions. 

%Moreover, FL also has limitations in accessing the full potential of available data due to reliance on static clients, leading to incomplete or biased dataset representations and affecting model performance. 

%In today's digital landscape, acquiring more clients is about efficiency. Docker Containers \cite{inton9} allow swift service deployment on fog/edge devices \cite{M7}, enabling high scalability. They also facilitate distributing ML models directly to clients, explored in our paper as On-Demand FL. Embracing containers for real-time deployment of learning algorithms unlocks new avenues for personalized services and adaptive decision-making.
%FL also faces challenges in adapting to data shifts \cite{kl3} and maximizing available data.

%FL poses challenges due to the diverse and dynamic nature of device availability and data characteristics. To address these challenges and optimize FL model deployment, 

%Advancements in mobile technology, exemplified by Apple's move towards on-device ML execution \cite{inton1}, signify a transformative leap for resource-constrained devices. This shift underscores devices' capability to independently execute complex ML tasks, reducing reliance on external servers.

As our clients become more available and distributed, it's important to implement a selection mechanism to guide the model learning effectively and address the challenges inherent in FL, as discussed earlier. Additionally, employing client selection enables us to optimize resource utilization by ensuring that only chosen clients participate in training. This strategy prevents unnecessary allocation of resources to unselected clients, thereby improving efficiency and maximizing the effectiveness of our On-Demand FL approach.
Dynamic device conditions require an adaptive approach for real-time deployment decisions. In today's varied applications, intelligent components are crucial for adaptive responses. Heuristic solutions for client selection \cite{M2}, %due to the NP-hardness of the problem, 
struggle to capture the changing nature of the FL environment. Relying on predefined rules, heuristics may not adapt well to rapid shifts in device availability.
The integration of Docker containers and Deep Reinforcement Learning (DRL) \cite{inton7} techniques has gained attention \cite{fixx1}. DRL outperforms traditional Reinforcement Learning (RL), particularly in scenarios involving tabular methods and linear approximations \cite{ahmed}. DRL's utilization of deep neural networks enables it to effectively handle high-dimensional state and action spaces, overcoming the limitations of tabular methods and linear approximations. By approximating complex and nonlinear functions, DRL captures intricate patterns and dependencies within the environment, facilitating better generalization across similar states and actions. Additionally, DRL enhances sample efficiency through techniques like experience replay and target networks, while its flexibility in network architecture and learning algorithms allows for adaptation to diverse problem domains. %Overall, DRL's ability to efficiently approximate complex functions, learn hierarchical representations, and adapt to real-world environments makes it a superior choice for reinforcement learning tasks compared to traditional methods.

Integrating client selection through DRL into On-Demand FL boosts adaptability in real-world scenarios, enhancing model convergence and efficiency. DRL's dynamic nature \cite{fixx4} optimizes Docker container placement, transforming deployment decisions based on real-time feedback. This flexibility ensures FL models are placed effectively, opening possibilities for future applications.
Deep Q-learning \cite{fixx3}, a key DRL component, excels at approximating complex functions, making it pivotal for tasks like resource management. Using deep Q-learning for client selection, model deployment, and demand analysis is promising. However, the exploration technique in RL may lead to initial errors, affecting convergence time.

This paper proposes a novel architecture that offers on-demand container deployment of models and learning mechanisms on clustered clients with orchestrators to manage them. It explores combining Docker containers and DRL for FL model deployment optimization, discussing technical aspects, advantages, challenges, and potential applications. The integration shows promise in enhancing FL convergence and performance in the dynamic environment, defining it as an MDP architecture \cite{inton6}.
Our MDP framework enables proactive decision-making on placement, adapting to FL application requirements and device environments. Our approach considers factors such as the quantity of active hosts, priority, data volume, and data quality. Using deep Q-learning in our DRL agent proves effective in intelligent decision-making amid environmental changes. Due to the exploration time challenges faced by DRL and the undesirable prolonged convergence in FL with additional rounds, we suggest a solution that merges offline and online learning, using a Master Learner and Joiner Learner setup. This includes deploying a mature decision-making model. By blending both approaches, we utilize historical data for initial model development and refine it in real time for adaptability to changing trends in dynamic environments.
Our approach outperforms heuristic, tabular RL, and the VanillaFL random \cite{vanilla} solutions tested on the MDC dataset \cite{dataa}. To our knowledge, this work is the first to address On-Demand client and model deployment in FL through DRL.

The contributions of this work are as follows:
\begin{itemize}
    \item We introduce a comprehensive end-to-end framework that leverages the On-Demand formation strategy, presenting our approach that circumvents the issues related to the extended learning duration inherent in DRL algorithms.

    \item We present a novel MDP framework designed to build a DRL agent. This framework adeptly considers factors such as resource availability and data shifts, enabling informed deployment decisions that align with the evolving environments of the devices.

    \item We utilize a custom DQN tailored to our problem domain to enhance learning precision and effectiveness of our novel On-Demand solution.

\end{itemize}

The rest of this paper is organized as follows. Section II represents the literature review. Section III describes the proposed architecture and methodology. In section IV, we illustrate the formulation of our model. Section V illustrates the experiments and results followed by a conclusion in Section VI.

\section{Literature Review}

The authors of \cite{M5}, introduce a framework designed for Internet of Things (IoT) applications that harness Docker containerization to address challenges and the limitations of underpowered fog devices. They present an On-Demand fog architecture that enhances device availability by efficiently deploying containers and installing required services on fog devices. Building on this, \cite{M7} and \cite{M2} extend the architecture further by proposing On-Demand micro-services and model deployment solutions that are cost-effective and maintain accessibility between end-users and available vehicular fog clusters, all facilitated by heuristic approaches.

In \cite{M8}, researchers delve into the complexities of training FL algorithms over real-world wireless networks. They tackle this challenge by formulating an optimization problem that considers both user selection and resource distribution to minimize the loss function. To address this, they derive a mathematical expression for the expected convergence rate of the FL algorithm, accounting for the influence of wireless factors. In another study, \cite{M9} introduces a decentralized FL approach at the segment level, aiming to improve network capacity utilization across client nodes. In addition to that, \cite{M10} adopts a combination of integer programming techniques to tackle resource allocation and deployment challenges in Mobile Edge Computing (MEC). In addition to that, in \cite{M11}, the authors explore the concept of expanding fog resources in terms of quantity and capability through the application of RL methods.

Furthermore, \cite{addkk3} presented a framework for FL, targeting non-IID data and limited network connectivity on mobile devices. By employing deep Q-learning for intelligent client selection, this approach minimized communication rounds while maximizing validation accuracy. Similarly, \cite{addkk4} focused on FL for heterogeneous and private IIoT data, leveraging DRL to select precise IIoT equipment nodes and enhance model aggregation rates, thereby reducing communication costs. Additionally, \cite{addkk5} utilized double deep Q-learning for client selection, considering trust and resource factors. Despite the efficacy of DRL in solving selection problems in FL, introducing new clients via containers amplifies the complexity of selection due to device dynamics, heterogeneity, and usage patterns. For that, a DRL solution to the On-Demand architecture necessitates the consideration of multiple additional objectives during client selection and deployment when compared to the previously mentioned DRL solutions.

%Moreover, The authors of \cite{addkk3} proposed a framework for FL, addressing non-IID data and limited network connectivity on mobile devices. By intelligently selecting client devices for FL rounds and optimizing device selection using deep Q-learning, this reduced communication rounds while maximizing validation accuracy. In addition to that, \cite{addkk4} focused on FL which was applied to heterogeneous and private IIoT data, with a focus on selecting accurate IIoT equipment nodes using DRL to enhance model aggregation rates and reduce communication costs. Similarly, \cite{addkk5} used double deep Q learning for client selection while taking into consideration trust and resource factors to select the client. Overall, DRL was used to solve the selection problem in FL. However, introducing new clients using containers renders the problem more complex in terms of selection due to the dynamicity and heterogeneity of devices and the nature of clients' usage. In addition to that, more objectives should be taken into consideration while selecting and deploying clients in the ON-demand architecture and the previously mentioned architecture.

In \cite{addkk2}, the authors developed a solution to address issues related to client availability, wherein models often exhibit bias towards clients with higher availability. However, this work did not propose a solution to maintain overall availability across multiple clients in diverse locations. Various strategies have been proposed in the literature to tackle data shifts in FL. Adaptive learning rate techniques and model personalization strategies have been explored to mitigate changes in data distributions among clients \cite{datashift1}. Additionally, methods such as domain adaptation and continual learning have been investigated to enable models to adapt effectively to shifting data over time \cite{datashift2}. While these solutions contribute to the resilience of FL models against data shifts, their primary focus has been on addressing distribution issues within available client data at specific stages. However, these approaches have not emphasized the concept of deploying clients strategically to ensure models learn from more diverse datasets. By deploying such clients, a larger set of clients becomes available for selection, along with access to more data.

\begin{figure*}[]
	\centering
	\includegraphics[width=0.8\textwidth]{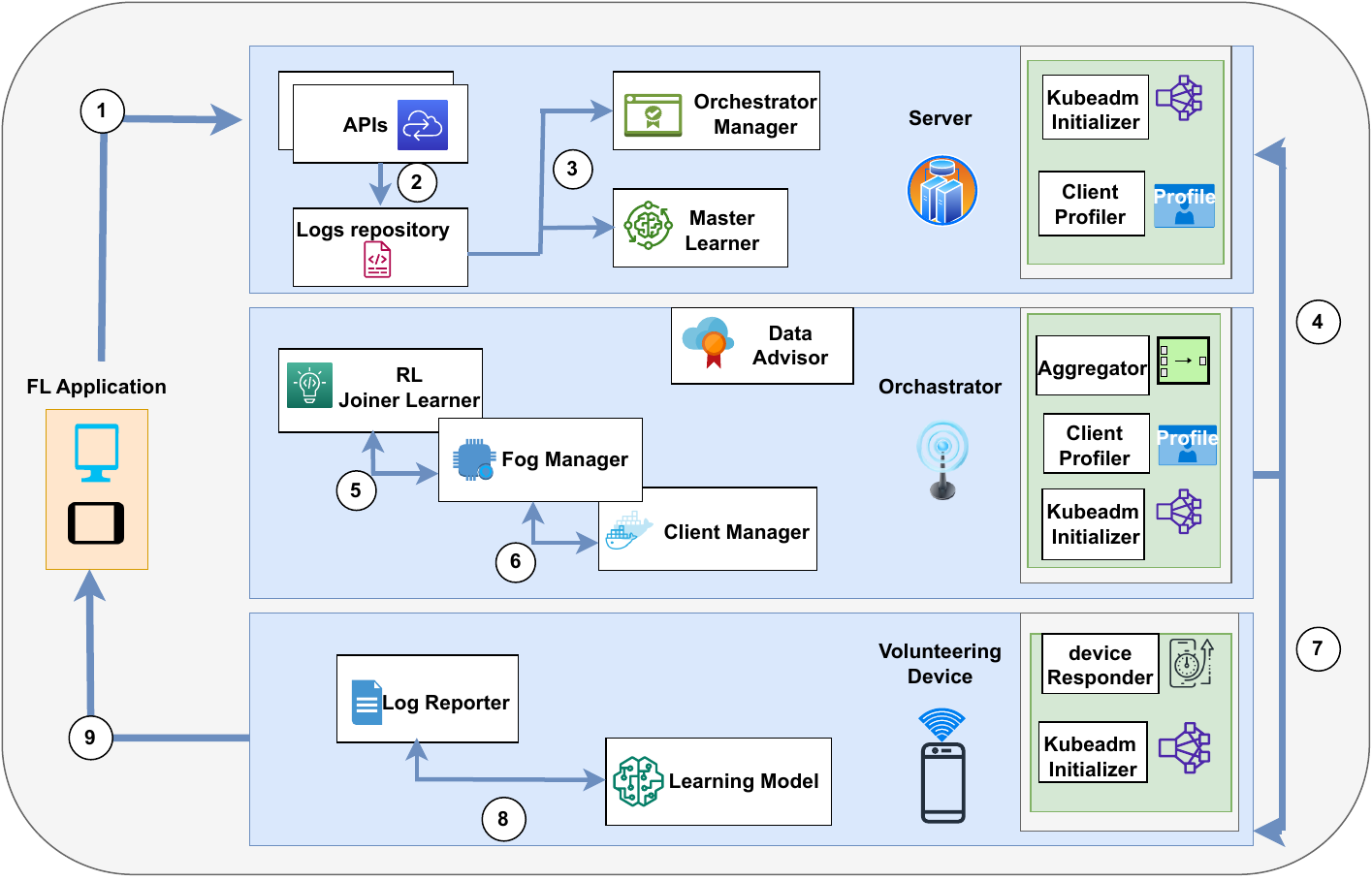}
	\caption{Overall Flow of The Proposed Architecture.}
	\label{fig2}
\end{figure*}

However, it's important to note that when there is a shortage of available clients for participation in FL, none of the client selection and optimization methods described earlier are applicable. Moreover, Adaptive learning rate techniques and model personalization strategies, though valuable, may struggle when confronted with scenarios where clients have intermittent or limited availability, potentially leading to incomplete model updates \cite{addkk2}. In addition, domain adaptation and continual learning methods to handle data shifts may not fully account for situations where new clients join or existing clients become unavailable during the learning process.
This paper places its focus on the challenge of ensuring available clients and the successful execution of FL applications in diverse scenarios where no prior research has explored the feasibility of employing a DRL solution for On-Demand FL. Given the advantages inherent in On-Demand FL deployment, which entails efficiently preparing and introducing clients to FL while optimizing their deployment and selection.
The issues mentioned above have captivated researchers' focus, prompting them to explore the utilization of DRL as a better alternative to heuristic methods. DRL is particularly advantageous as it considers the shifts in data availability and accommodates changes, offering a more effective approach to address these challenges. This motivation led to the proposal of a DRL solution for the On-Demand FL model and client deployment architecture.
To the best of our knowledge, there has been no previous work that has provided the capability to instantaneously deploy FL clients using DRL whenever and wherever volunteer devices are accessible but don't have the required software to train and download the model, which in turn enables the environment to engage in model training and leverage the resulting data effectively.

\section{Proposed Architecture}

The adoption of FL is important because of the paramount concern for data owners' privacy, driving the necessity for a decentralized approach that seamlessly integrates intelligence while safeguarding sensitive information. FL revolutionizes the traditional model by enabling the collaborative training of a global model distributed across various clients, leveraging their localized data. Through iterative aggregation of updated parameters from individual client training sessions, the server orchestrates the refinement of the global model, thereby enhancing convergence and accuracy over successive rounds \cite{kl5}. However, client failures or delayed parameter reports pose a significant challenge, potentially leading to excluding learning rounds and disrupting the learning process. Thus, while selecting clients, ensuring their consistent availability and substantial resource allocation for active participation is essential. Additionally, creating a learning environment involves establishing a connection, gathering resources, installing components, and starting the training protocol.

The proposed architecture, as shown in Figure \ref{fig2}, consists of three distinct entities: the server, orchestrator, and volunteering device, each entrusted with specific responsibilities. The server assumes the centralized management of client provisioning, the global model, and secure connections, while orchestrators administer Kubeadm clusters \cite{addk1}, supervise container deployment, monitor device dynamics, and integrate new clients while handling the deployment and selection process in their area of reach. The reason for selecting a Kubeadm cluster is its ability to swiftly generate and set up clusters without the necessity of manually dealing with the cluster configuration. This tool simplifies the procedure by automating tasks like initializing the control plane, integrating nodes into the cluster, and setting up network plugins. In instances of significant device mobility, orchestrators seamlessly manage client requests, forwarding them to the server for container deployment, thus fortifying the system against single-point failures and mitigating server overload. Each orchestrator operates within a defined area, managing a finite set of clients within its reach and availability. These clients are the ones accessible within their designated geographical area, ensuring that the orchestrator efficiently handles a manageable number of connections. Through the utilization of lightweight containers and the inherent flexibility of the architecture, clients can execute ML models from any location, thereby ensuring optimal resource efficiency. Operating exclusively on devices engaged in FL rounds, clients can run ML models from any location, we maximize resource utilization and enhance the efficacy of the learning process.

The process depicted in Figure \ref{fig2} initiates with a FL application triggering the server to deploy devices strategically positioned within reachable areas to serve as FL clients. These applications maintain comprehensive logs tracking user connections, IP addresses, ML model requests, and device movements, which are instrumental for the orchestrator manager in orchestrating orchestrators and invoking On-Demand client deployment as needed.

\subsection{Joiner Learning}
This component processes a deep analysis and model refinement for a well-established DRL model. Furthermore, this component operates online, offering adaptability to changing data and conditions. By continuously monitoring the latest data inputs and environmental factors, the model can dynamically adjust its behavior and decision-making processes. This real-time adaptability ensures that the mature model remains relevant and effective in response to evolving scenarios, thereby maximizing its utility and performance over time.

\subsection{Master Learner}
Utilizing the offline DRL algorithm offers a significant advantage in expediting deployment decisions within the system \cite{kl6}. By processing and learning from this historical data, the offline DRL algorithms gain valuable insights into optimal deployment strategies and patterns. This enables the system to make informed decisions quickly, without the need for extensive trial-and-error experimentation. Subsequently, the DRL Joiner Learner component plays a crucial role in enhancing the efficiency of the learning process. Rather than starting from scratch with each new model iteration, the DRL Joiner Learner receives a mature model, from this component, that has already undergone significant refinement and optimization. This approach circumvents the need for long learning periods typically associated with training new models from scratch. Instead, the mature model serves as a solid foundation, facilitating rapid integration into the system and contributing to the creation of an On-Demand learning environment.

\subsection{Kubeadm Environment Initialization}
On available fog devices in each area, the server begins by deploying a few orchestrators based on recent client device motion data. These orchestrators oversee the setup process. In areas with many willing volunteer devices, orchestrator deployment is prioritized. The server considers weekday/weekend status for orchestrator placement. Assuming a list of potential mini-server orchestrators, the server identifies Kubernetes cluster locations and enrolls volunteer devices during orchestrator node configuration. All eligible devices in the area join the Kubeadm cluster \cite{kl7}. With the environment prepared, services can be deployed promptly. It's crucial for the Master node to remain active, as the cluster fails if it goes down. If the initial orchestrator becomes unavailable, another mini-server is promptly selected to prevent restarting the Kubeadm cluster. It's presumed each region has enough mini-servers, but if not, a client device may temporarily act as an orchestrator based on its profile. 
The server triggers the Kubeadm initializer for instant cluster creation and environment preparation. The Kubeadm initializer manages cluster health and device and model conditions. Models are deployed, and clients are selected based on the DRL Joiner Learner's output.

\subsection{Fog Manager}

This component plays an important role in overseeing the activity of nodes within the system, particularly focusing on two key aspects: node time and stay duration in specific regions. By monitoring node time, it tracks the duration for which each node remains active and engaged within the system. This information provides valuable insights into the level of participation and contribution from individual nodes, helping to gauge their overall involvement in the training process.
Additionally, the component monitors stay duration in different regions, which refers to how long nodes remain within particular geographical areas. This data allows for a deeper understanding of node mobility patterns and the distribution of resources across different regions. By analyzing stay durations, the system can identify areas with high node activity and prioritize resource allocation and training efforts accordingly.
Furthermore, this component actively contributes to achieving training goals by ensuring that nodes are effectively utilized and coordinated.

\subsection{Client Manager}
This component is responsible for managing the mobility of devices within the system, primarily focusing on detecting movements and facilitating communication with the server for container deployment and client selection processes. By continuously monitoring the movements of devices, this component ensures that the system remains responsive to changes in device locations and can efficiently deploy containers and select clients accordingly.
Furthermore, mobility generates valuable data that can be leveraged to enhance model training. By tracking the movements of devices, the system gains insights into spatial dynamics and user behaviors, which can be utilized to improve the accuracy and robustness of our model. For example, patterns in device movement may reveal trends or correlations in user behavior that can change the model learning trends and lead to more accurate predictions or classifications. It also tracks user participation rounds to prevent selection bias.

\subsection{Data Advisor}
%This component accesses the global accuracies of each round and monitors the movement patterns of clients, it detects variations in data availability within specific geographical areas. Sending insightful feedback to the orchestrator contributes to the adaptation of the DRL agent to these changes and the dynamic shifts of data from one area to another. This iterative process aids in refining the learning model, ensuring its responsiveness to evolving data distributions and enhancing the overall efficiency of the FL system.

This component serves as a crucial analytical engine within the system, tasked with two primary functions: accessing the global accuracies of each training round and monitoring the movement patterns of clients. By accessing global accuracies, it gains insight into the overall performance and effectiveness of the ML model across various rounds of training. This information provides valuable feedback on the system's progress and helps identify trends or patterns in model performance over time.
Simultaneously, the component diligently tracks the movement patterns of clients, enabling it to detect variations in data availability within specific geographical areas. By monitoring changes in data availability, it can identify regions with fluctuations in data quality or quantity, as well as areas experiencing dynamic shifts in data distributions.
The component then leverages this information to provide insightful feedback to the orchestrator, contributing to the adaptation of the DRL agent to these changes and dynamic shifts in data. This feedback loop enables the system to dynamically adjust its strategies and resource allocations in response to evolving data distributions and client movements.

\subsection{Aggregator}
The aggregator plays a crucial role in refining the global model through the application of an FL aggregation function, all while monitoring the weight updates during each round \cite{kl1}. Additionally, it's tasked with assessing the significance of each round, ensuring its validity by counting the number of received weight updates. Should the count of updated weights fall below a predetermined threshold established by the server, the aggregator takes proactive measures to omit the round, thereby maintaining the integrity and efficiency of the FL process.

\subsection{Orchastrator Manager}
This component records client interactions with chosen orchestrators, providing insights into user behavior within specific locations. While selecting nodes as orchestrators raises security concerns, this aspect falls beyond the scope of this study, assuming the server's confidence in accessible mini-servers.
To assist in orchestrator selection, historical user data on model training is utilized. Client movements are factored in when choosing orchestrators, with considerations such as weekend versus weekday deployments. For instance, orchestrators may be deployed in areas with mountainous terrain during weekends when user activity is anticipated to be higher.
This process occurs at regular intervals, with each orchestrator subsequently tracking client movements. Furthermore, the module manages requests from orchestrators to deploy additional containers in their respective areas.

\subsection{Client Profiler}
This component collects device data, including battery life, resources, and location history. Monitoring these variables enables the system to optimize resource allocation and workload distribution.

\subsection{Log Reporter}
This component is responsible for tracking performance metrics and updating client logs based on the output generated by the FL model. This entails recording various performance indicators such as accuracy, loss, convergence metrics, selected clients, and other relevant statistics derived from model training and evaluation processes.

\subsection{Learning model} 
This model is the ML algorithm deployed within the FL framework for training on client devices. It's adaptable to different FL applications, varying based on data characteristics and system objectives. Its design and parameters are tailored to handle diverse data and meet specific system goals.

\subsection{Log Repository} 
The repository serves as a comprehensive archive containing valuable information regarding previous deployment strategies, algorithms employed, parameters utilized, and combinations of states and actions. This repository proves invaluable for any ML model or agent seeking to leverage historical data as a foundation for learning and improvement. By relying on this repository, ML models can have insights from past experiences and utilize them as a basis for offline learning and model enhancement.

\subsection{Device Responder} 
This component facilitates the communication between the container hosted on the device and the device\ fog itself. Its primary responsibility is to respond promptly to any requests initiated by the container, enabling access to the data required for learning or deployment purposes.
In essence, this component acts as a gateway, ensuring that the container has easy access to the necessary datasets, models, and the resources of the devices.

\section{Formulation and Solution}
\subsection{Problem Modeling}

MDP is a mathematical framework for sequential decision-making in uncertain environments, represented as $(S, A, Pr, C, \gamma)$. Here, $S$ stands for possible agent states, $A$ is the action space, $Pr$ signifies transition probabilities, $C$ is the cost function, and $\gamma$ determines the weight of future rewards. To apply MDP to our problem, we must define these elements: $S$, $A$, $Pr$, $C$, and $\gamma$.

The RL agent's decision occurs at round $r$ within the range of $R = [1, 2, . . . , \mathcal{R}]$. Let $F = [F_1, F_2, . . . , F_m]$ represent the list of $m$ devices, where each device is described by attributes like CPU, memory, diskspace, availability, and area location, denoted as $F_i = [F_{i,cpu} , F_{i, memory}, F_{i, diskspace} , F_{i, availability}, F_{i, area}]$.

Similarly, let $P = [P_1, P_2, . . . , P_n]$ denote the set of $n$ models for placement. These models differ in their resource requirements and have priority values, indicating the need for prioritized placement to maintain FL application accuracy. A model $P_i$ is defined by requirements such as CPU, memory, diskspace, and priority, represented as $P_i = [P_{i, cpu} , P_{i, memory}, P_{i, diskspace} , P_{i, priority} ]$. 
Each device varies in its resource utilization for the same model, depending on the specific task and data volume it handles.

The placement decision is made step by step, considering each model container, %within a state at time $t$
until all available clients have been examined. The overall placement decision, represented as $k$ in round $r$, is a one-dimensional array of size $n$. In this array, the value $k_{i}$ = 1 indicates that model $P_i$ is deployed on device $F_i$ %at time $t$
during round $r$, while 0 signifies otherwise.

The available actions consist of two options: (1) selecting a client to deploy the model on, or (2) excluding a client from the round. When a device like $F_i$ is chosen, the action involves placing model $P_i$ on that specific device. Conversely, if the client is not selected, the model remains undeployed. For that, the action of each round $r$ is defined as $a=\{t_1,...,t_n\}$,
%This action set $A$ can be represented as  $A = [0, 1]$,
and we can denote a typical action element at client $t$ during a round $r$ as $a_{t}$. During every round of FL, the server chooses participants from the total pool of $n$ available clients which is $k$. %Consequently, the action taken in each round, denoted as at, consists of selecting a group of clients represented

Our state space is represented by $s \in S$, with $s = (k, AC, r, DA)$. Here, $AC$ signifies the enhancement rate of the global accuracy of the model, $r$ indicates the round, and $DA$ represents the requested area for deploying clients by the orchestrators.

%Furthermore, a counter u is used to indicate the current container Pu the agent is taking the decision for, such that $u \in [1, . . . , n]$. After making n decisions, the counter is reset to one. Henceforth, the state of our model is:

\subsection{Cost Function}

The cost function considers four conflicting objectives: (1) Minimizing deployed clients, (2) Maximizing data volume and diversity, (3) Minimizing the number of unplaced high-priority models, and (4) Maximizing the orchestrator's requests for model deployment. We introduce a cost function that combines these objectives, with adjustable weights $W$ in the range of $[0, 1]$. These weights allow flexibility in reflecting the FL application's preferences, where higher weights indicate greater impact on the objectives. The weights sum up to 1.

The first goal is to minimize the number of active hosts, conserving energy and resources like battery, CPU, memory, and disk-space \cite{kl8}. This boosts the availability of fog devices, allowing other applications to utilize the extra hosts in the area. Additionally, it reduces network congestion in FL by reducing the high rate of parameter exchanges and updates.

\begin{equation}
    c_1= w_1 \times \sum_{i=1}^{n} k_{i}
\end{equation}
%Here, $k(r)$ is a one-dimensional binary list of size m. 
If $k_{i}$ = 1, it signifies that the fog device $F_i$ is hosting a model.

The second objective aims to enhance both the quantity and diversity of data. In ML, a larger and more diverse dataset typically leads to improved model performance \cite{kl9}. Selecting clients with high mobility from various areas generates a substantial volume of data, ultimately contributing to better model performance.
\begin{equation}
    c_2= w_2 \times ER
\end{equation}

Here, $ER$ denotes the variation rate, which quantifies the diversity in area locations within each set of selected highly mobile clients. This value is calculated by checking if the area locations that the selected clients were chosen are different, and if they are flagged as high movements clients or not. A low $ER$ score indicates that the chosen clients come from different areas and exhibit significant movement. Data produced by these clients is diverse, particularly in terms of output class features. We're not intentionally making the data less evenly distributed (non-IID), but deploying more clients increases data volume. Non-IID data in FL refers to diverse data distributions across clients, challenging the aggregation for a global model while maintaining privacy and data locality. This objective allows more diverse class labels available within each area for learning, and shared output class labels available across different area locations for a broader understanding.

The third objective is to reduce the count of high-priority clients that remain undeployed. High-priority clients possess high local accuracies, indicating their data strongly correlates with the model \cite{kl10}. Deploying models on these clients at early rounds results in achieving high accuracy for both the models and FL applications.

\begin{equation}
    c_3= w_3 \times max(\sum_{i=1}^{n} k_{i} \times P_{i, priority})
\end{equation}

%Here, $fg(t)$ is a one-dimensional binary list with a size of $n$. If $Fg_j (t)$ = 1, it indicates that the jth model has been deployed on a device in the set $F$, and 0 otherwise. The vector $1 - fg(t)$ results in a value of 1 if the model is not placed on any client.

The fourth objective is to maximize fulfilling orchestrator requests for deploying model containers in high-activity areas. Orchestrators monitor user movements and locations, and when there's significant activity in an area, a message is transmitted to the server, asking for the deployment of containers on clients situated in those areas. Furthermore, the data advisor component plays a crucial role in achieving this objective by providing feedback to the orchestrator regarding the drift in data availability in specific geographical areas. This proactive engagement enhances the overall quality of the learning process, concurrently minimizing the need for discarding rounds.

\begin{equation}
    c_4= w_4 \times max(\sum_{i=1}^{n} k_{i} \times RT )  
\end{equation}

Here, $RT$ represents the mean rate of clients chosen from the designated areas.

%\begin{equation}
%     Pun_{memory} =  \sum_{i=1}^{m}  max (\sum_{j=1}^{n} (P_{j_{cpu}} \times K(t)_{ij}) - F_{i_{cpu}} , 0 )
%\end{equation}
   
%The same logic is applied to the other resources. $Pun$ = $Pun_{memory}$ + $Pun_{cpu}$
%+ $Pun_{hardisk}$ + $Pun_{battery}$
To penalize the agent for generating an impractical solution, we apply a punitive approach denoted as $Pun$ with a default value of 1.

For each client $F_{i}$, there is a utilization factor $P_{i}$ that signifies the extent to which a service consumes the resources of that specific client. This constraint is mathematically defined as follows:
    
    \begin{equation}
    P_{i,CPU} \times k_{i} \le F_{i,CPU}
    \end{equation}
    \begin{equation}
    P_{i,memory} \times k_{i} \le F_{i,memory} 
    %U_{memory_i} \le K_{memory_i}
    \end{equation}
    \begin{equation}
    P_{i,diskspace} \times k_{i} \le F_{i,diskspace}
    %U_{diskspace_i} \le K_{diskspace_i}
    \end{equation}
    \begin{equation}
    P_{i,Battery} \times k_{i} \le F_{i,Battery}
    %U_{Battery_i} \le K_{Battery_i}
    \end{equation}
    
    $\forall i \in \{1, \dots, n\}$ \\%i.e. for all available clients $F_{i}$ and their utilization $P_{i}$.\\
Any deviation from these rules results in increasing the value of $Pun$ by 1; otherwise, it adds 0.

In addition, some clients might not be available for the entire round, so the selected client must be accessible for a time greater than the average round duration to prevent interruptions due to client mobility. The server sets a parameter ${T\in N}$ as the minimum round time required. Thus, a client can be chosen and deployed if $F_{availability_i}$ (the time $F_{i}$ remains in its area) is greater than $T$.
    
    \begin{equation}
    \forall i \in \{1, \dots, n\} \;\;\;\;\; F_{availability_i} \ge T
    \end{equation}

Any deviation from these rules results in increasing the value of $Pun$ by 1; otherwise, it adds 0.

The overall cost is the aggregate of various cost measures and constraints, which are applied as penalty methods.
\begin{equation}
    C(s,a) = (c_1+c_2-c_3-c_4) \times Pun
\end{equation}

In our scenario, the minimization of function $C$ corresponds to the optimization of the deployment and selection problem. The ideal situation is when $Pun$ equals 1, indicating that no violations have occurred.

\subsection{Solution}

In our FL algorithm, %outlined in Algorithm \ref{alg:cap1}
the orchestrator employs the Joiner Learner Algorithm to deploy clients as needed, initiates the FL process, and updates the global model accuracy, as well as the Master Learner model, with information about the current round and update the model.

%\begin{algorithm}

%caption{Joiner Learner Algorithm}\label{alg:cap1}
% \textbf{Data:} Initialize the global parameter set of the base ML model

%\For{\textbf{each} round r =1, 2, 3.., r}  
%\begin{enumerate}
%\item Use Master Learner model to select and deploy clients
%\item Send the updated weights to the aggregator
%\item Updated local and global weights
%\item Update Master Learner model
%\end{enumerate}
%\EndFor

%\State\textbf{Return:} Return final models to FL application 

%\end{algorithm}

Moreover, %as shown in Algorithm \ref{alg:cap2},
to enhance our policy without relying on a model of the environment, the agent needs to understand the value function. In our specific context, we lack knowledge of the environmental dynamics. An alternative approach involves working with the state-action value function, $Q(s,a)$, which allows us In this scenario to minimize $Q(s,a)$ and discover the optimal minimum value.

\begin{algorithm}
\SetAlgoLined

\caption{DRL Algorithm using DQN for client selection in On-Demand FL}\label{alg:cap2}

 %\begin{itemize}
    Initialize Q-network with random weights\;
    Initialize target Q-network with the same weights\;
    Initialize replaybuffer, $\theta$, and $\theta$'\;
    $\theta$' = $\theta$\;
 %\end{itemize}

\For{\textbf{each} episode }  {
%\begin{enumerate}
    Initiate FL model weights\;
    initialize $s = (K, AC, r, DA)$ to zeros\;
  
%\end{enumerate}

\For{\textbf{each} $r$ in episode}{
%\begin{itemize}
    Load the $Q$ values by feeding the current $s$ to the Q-network\;
    Use the Top-p sampling strategy for $Q$ to select the clients and obtain the action $a$ at round $r$\;
    %\item Choose action $a$ based on exploration-exploitation trade-off:
    %\begin{enumerate}
    %    \item $A$=RandomAction() with probability $\epsilon$  OR
    %    \item $A$=argmax(Q-network.predict(s)) with probability $1-\epsilon$

    %\end{enumerate}
    Calculate reward/cost $C(s,a)$\;
    Store transition $(s,a,C(s,a),s')$ in replaybuffer\;
    Sample mini-batch from replaybuffer\;
    
   % \begin{itemize}
    \For{\textbf{each} $i$ in length(minibatch)}{
        %\begin{itemize}
            $target = C_{i} + \gamma  min(Q(s'_{i},a'_{i},\theta')) $\;
        %\end{itemize}
    %\EndFor
   % \end{itemize}
    }
    Update $\theta$ towards minimizing the loss using gradient descent $(target - Q(s,a,\theta))^2$  \;
    Periodically update target Q-network weights $\theta$' $\leftarrow$ $ \theta $ \;
    
    $s$=$s'$ \;
%\end{itemize}
}
%\EndFor
    
}
%\EndFor

%\textbf{Return:} Return Model

\end{algorithm}

The RL agent evaluates its performance using a state-action cost function, denoted as $Q(s, a)$, computed within a matrix or Q-table with dimensions $S \times A$. Typically initialized with zeros, this matrix aims to find the optimal Q-function, updating Q-values at each round $r$.
To enhance global model accuracy and consider round number, the agent incorporates these factors into the state. This enables monitoring of the global model's performance, facilitating the identification of clients causing performance drifts to avoid their future selection.
As shown in Algorithm \ref{alg:cap2}, Deep Q Learning extends Q-learning by integrating a Neural Network, surpassing traditional methods. Instead of relying on a Q-table, this approach utilizes a Neural Network to process a state, estimating Q-values for each action. This shift addresses the limitations of Q-tables in scenarios with numerous actions and complex game states, where traditional methods prove impractical, inefficient, and slow. The integration of a Deep Neural Network allows for a more scalable solution. The network receives the state as input and generates distinct Q-values for each available action. The ultimate decision involves selecting the action associated with the best Q-value. While the fundamental learning process remains unchanged, employing an iterative update approach, the key divergence lies in the mechanism of improvement. Instead of updating a Q-table, the weights in the neural network are adjusted, enhancing the quality of the outputs and ensuring adaptability. Deep Q-Network (DQN) uses experience replay, where past experiences are stored in a replay buffer and randomly sampled during training. This helps break correlations between consecutive samples and leads to more stable learning. Moreover, it employs two separate neural networks. The target network's parameters are periodically updated with the main network's parameters, providing more stable targets for the Q-values. The agent uses the top-p sampling strategy \cite{topp} to choose a group of clients associated with the action $a$ to take part in a round $r$ of FL. The idea behind top-p is that rather than examining every possible action, the agent arranges actions according to their probabilities, such as the softmax probabilities derived from Q-values. It then picks a subset of actions that together represent a specific proportion, $p$, of the probability distribution. This method ensures that the agent explores a variety of actions while giving priority to those with higher probabilities of being the best choice. This way, the agent can balance between exploiting well-known, high-value actions and exploring potentially beneficial but less common actions.

The overall Q-value formula for DQN is derived from the Bellman equation and is represented as follows:
    \begin{equation}
    \begin{split}
        Q(s,a;\theta) = (1-\alpha). Q(s,a, \theta) + \alpha (C(s,a) \\ + \gamma \times minQ(s',a';\theta '))    
    \end{split}
    \end{equation}
where $\theta$ represents the weights of the neural network. The symbol $\alpha$ signifies the learning rate, while $s'$ and $a'$ stand for the subsequent state and action. $minQ(s',a';\theta ')$ represent the minimum Q-value for the next state $s'$ over all possible actions $a'$ according to the target network with parameter $\theta'$

Afterward, the FL scenario is shown in Algorithm \ref{alg:cap1}. Here, the server distributes the global model weight parameters to the clients. Subsequently, the algorithm employs Algorithm \ref{alg:cap2} to designate a subset of clients to partake in the ongoing FL round. These chosen clients then proceed with their learning process, transmitting their updated weights back to the server. Ultimately, the aggregator leverages the FedAvg algorithm to revise the parameters of the global model based on the received updates.

\begin{algorithm}

\caption{On-Demand FL Algorithm}\label{alg:cap1}
 \textbf{Data:} Initialize the global parameter set of the base ML model and distribute them over the clients\;

\For{\textbf{each} round $r =1, 2, 3.., \mathcal{R}$ }  {
%\begin{enumerate}
Use Algorithm \ref{alg:cap2} to select and deploy clients\;
Perform local training\;
Send the updated ML weights to the aggregator\;
Update local and global weights\;
%\item Update Master Learner model
%\end{enumerate}
%\EndFor
}
\textbf{Return:} Return final models to FL application\;

\end{algorithm}

%Within the equation of $Q(s, a, \theta)$, The symbol $\alpha$ signifies the learning rate, while $s'$ and $a'$ stand for the subsequent state and action.
%\begin{equation}
    %Q(s,a) = Q(s,a) + \alpha (C(s,a)+ \gamma \times %MinQ(s',a') - Q(s,a))
%\end{equation}

\section{Results and Analysis}

\begin{figure*}[htp]
	\centering
	\includegraphics[width=0.9\textwidth]{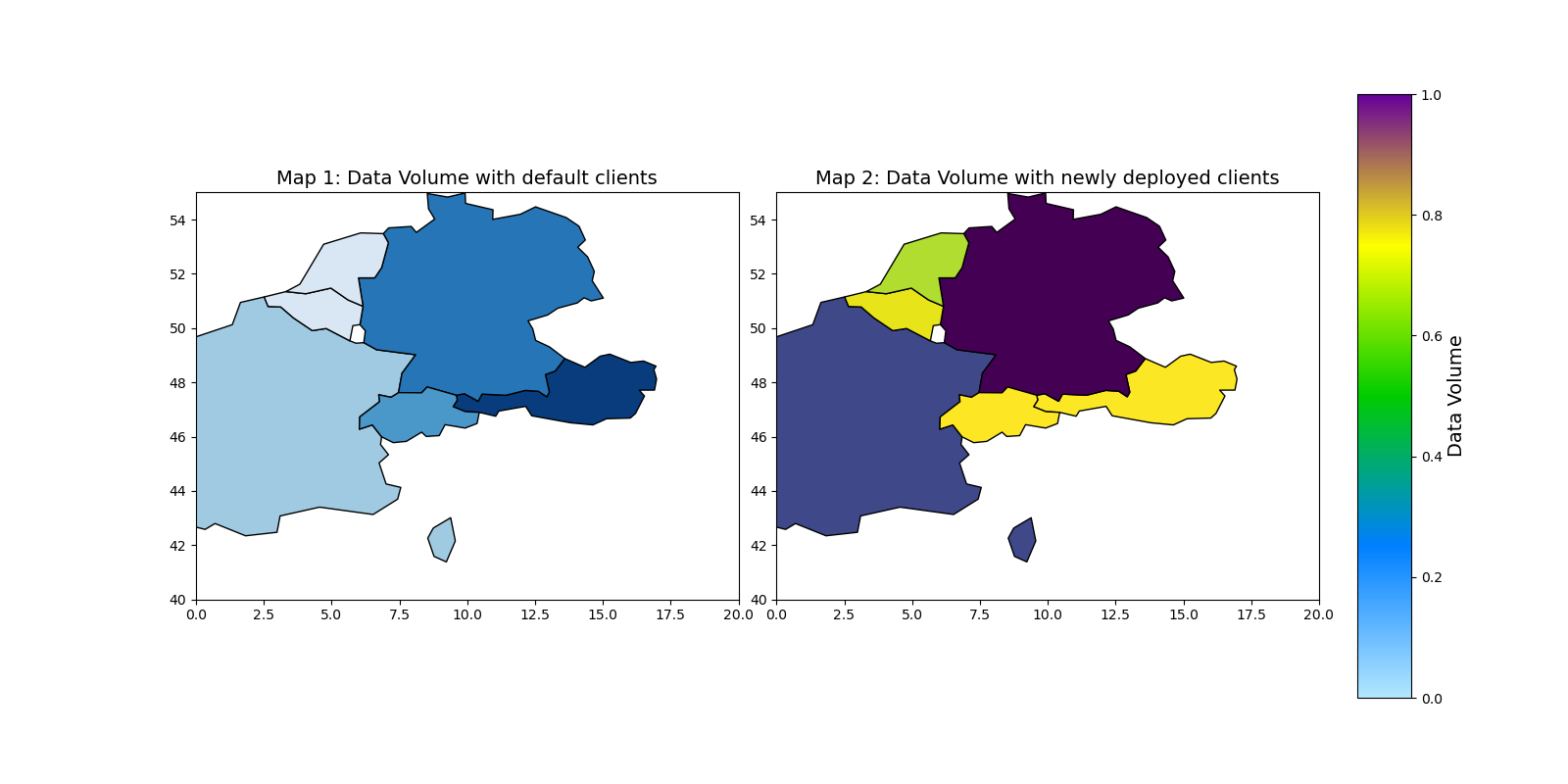}
	\caption{An instantaneous view of the data volume of records accessible for learning, comparing the default clients with those dynamically deployed On-Demand.}
	\label{fig12}
\end{figure*}

\begin{figure}[htp]
	\centering
	\includegraphics[width=0.45\textwidth]{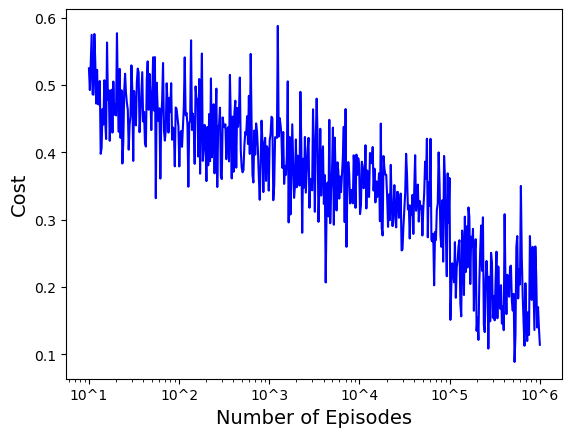}
	\caption{Cost Evolution Over Time}
	\label{fig5}
\end{figure}

\subsection{Setup} We opted for the utilization of the MOBILE DATA CHALLENGE (MDC) dataset \cite{dataa} in our learning process due to its unique relevance to geographical locations. This dataset provides a comprehensive collection of continuous client movement records, enabling us to extract features that are important in predicting clients' subsequent locations. The inclusion of place details, time indicators, and destination labels in the dataset allows our model to showcase the real-world significance of On-Demand FL. By working with this dataset, we aim to demonstrate the applicability of On-Demand FL in scenarios where client movements in geographical locations play a pivotal role, emphasizing its importance in real settings.

To facilitate FL, we used the ModularFed framework \cite{arafehlocalfedd}, which is known for its FL application capabilities. Before sharing a global model with clients, we first built centralized models for a specific application. For the Next Place Prediction, we optimized a Deep Neural Network %\cite{neuraln}
with three
hidden layers of size 128 and 256, and ”Relu” and ”softmax” as activation functions. We got an accuracy of 60\%.

Our dataset contains 47 users, and clients were assigned data records associated with a single user. The data distribution across clients was Non-iid. %In each learning round, the server decides the fraction of deployed clients $L$.
We set the learning rate $\alpha$ to 0.01, and the discount factor $\gamma$ to 0.9. The source and target networks of our DRL are four layers of deep neural networks with input and output of size $n$ and hidden layers of size 128, 256, and 128 respectively with Adam optimizer. %We used ReLu activation functions for the hidden layers and a linear activation function for the output layer. 
The threshold probability $p$ for the top-p sampling strategy is configured to be 0.9. The learning phase underwent 5 iterations before obtaining the results.

\subsection{Experiments}

In this section, we compare our approach with heuristic implementations based on genetic algorithm, tabular RL, and the default random method of selecting clients. Our contribution is the first to address On-Demand FL through DRL, and our thorough comparative analyses distinguish our work. We are not comparing with other literature works that adapt DRL for client selection in FL since our On-Demand context requires different objectives.
In the following sections, we'll demonstrate how our solution facilitated the DRL agent's learning progress over time, along with showcasing cost variations. Additionally, we'll present statistics indicating an increase in the number of participating clients in FL rounds compared to previous static deployments and how that helps in managing data shifts and availability in the learning area locations. Moreover, we'll compare the overall accuracy levels achieved by various implemented solutions across different scenarios. Finally, we'll assess how our solution effectively maintains high accuracy levels while minimizing the number of discarded rounds during the learning phase.

%\begin{enumerate}

1) Figure \ref{fig5} illustrates the average cost evolution over time. 
The graph is presented using a logarithmic scale to enhance the clarity of visualization for the agent's performance. It is evident that our solution successfully converged and reduced the cost function while adhering to the predefined objective functions and constraints. The convergence and scalability are compatible with the number and diversity of clients we have. %These include minimizing device usage, maximizing data volume and diversity, and maximizing service requests to deploy clients in specific areas. 
This convergence process contributes to the development of the intelligent Master Learner component. Subsequently, this pre-trained DRL model through our offline mechanism is used by orchestrators for the On-Demand deployment of clients and models, all without any delays. At this stage, the orchestrator seamlessly utilizes this mature model to initiate the FL application.

%\begin{table}
%  \centering
%  \caption{An instantaneous view of the data volume of records accessible for learning, comparing the default clients with those dynamically deployed On-Demand.}
%  \begin{tabular}{|p{2cm}|p{2cm}|p{2cm}|}
%    \hline
%    & Default/Static & On-Demand \\
%    \hline
 %   Area 1 & 1000 & 4500\\
 %   \hline
 %   Area 2 & 2050 & 6070 \\
 %   \hline
  %  Area 3 & 900 & 4200\\
 %   \hline
 %   Area 4 &  2650 & 10105 \\
 %   \hline
 %   Area 5 & 4210 & 11420\\
 %  \hline
 %   Area 6 & 3870 & 10360 \\
 %   \hline
 % \end{tabular}
  
 % \label{tab:your_table_label}
%\end{table}

2) Our innovative approach to On-Demand client deployment has proven to be a game-changer in the world of FL, significantly enhancing client engagement and data availability. By seamlessly integrating our methodology, we have observed a notable number of clients actively participating in the learning process. This act, as reflected in Figure \ref{fig12}, underscores the efficacy of our strategy in not only attracting a higher number of clients but also fostering sustained engagement throughout the learning period. The noticeable increase in data availability is readily apparent when introducing new clients, characterized by deeper and more vibrant colors, in sharp contrast to the relatively static presentation observed when relying on fixed clients. The consequential increase in the volume of available data is invaluable for FL, as it directly translates to a richer and more diverse dataset for model training. The positive correlation between our On-Demand deployment and enhanced client participation highlights the important role our approach plays in optimizing the FL ecosystem, ultimately contributing to more robust and accurate ML models. At this point, the orchestrators have a greater quantity of clients available for selection, and the process of selection will be further examined in the upcoming sections.

Furthermore, in instances where the FL environment encounters data shifts due to changes in client locations, resulting in decreased data quantity and diversity in certain areas, our solution assumes a critical role in effectively addressing these shifts. Specifically, leveraging the fourth objective function, our approach empowers orchestrators to strategically determine the deployment of containers and the establishment of new clients in regions experiencing data scarcity and a shortage of available clients due to these dynamic data shifts.

\begin{figure}[] 
	\centering
	\includegraphics[width=0.45\textwidth]{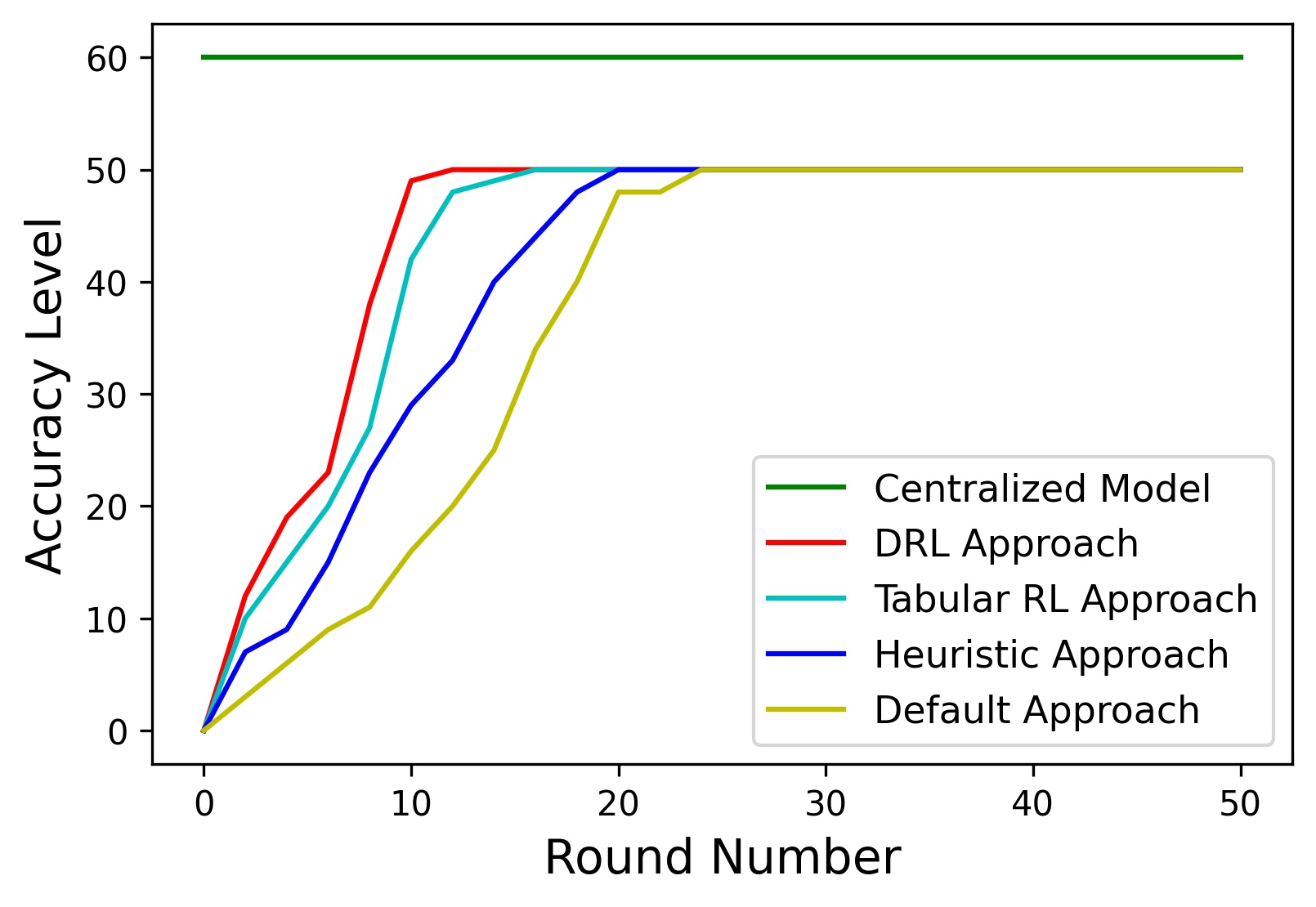}
	\caption{Accuracy with Relation to the Number of Rounds}
	\label{fig4}
\end{figure}

\begin{figure*}
\begin{subfigure}{.5\textwidth}
  \centering
  \includegraphics[width=.78\linewidth]{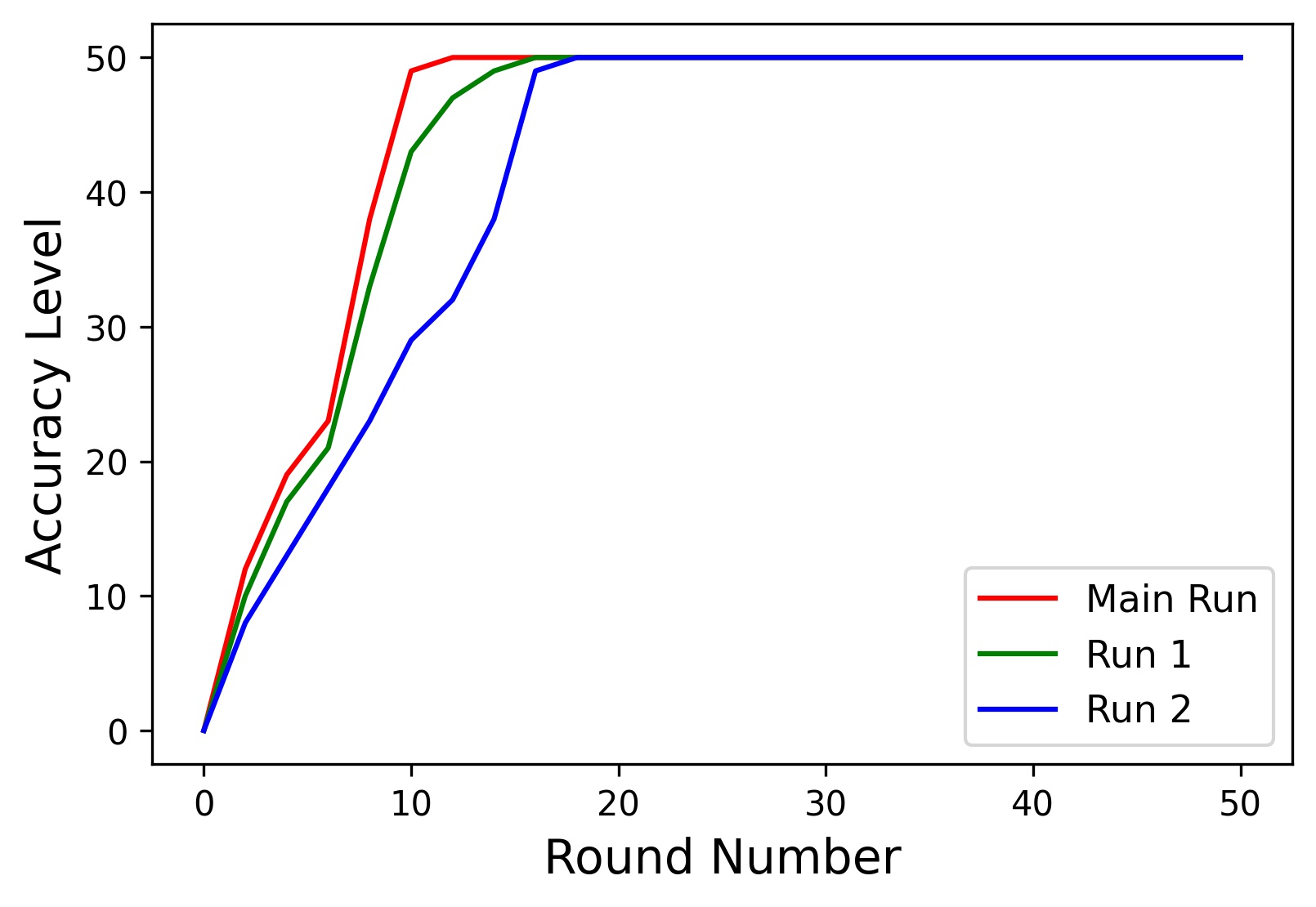}
  \caption{6.a}
  \label{fig:sfig1}
\end{subfigure}%
\begin{subfigure}{.5\textwidth}
  \centering
  \includegraphics[width=1.0\linewidth]{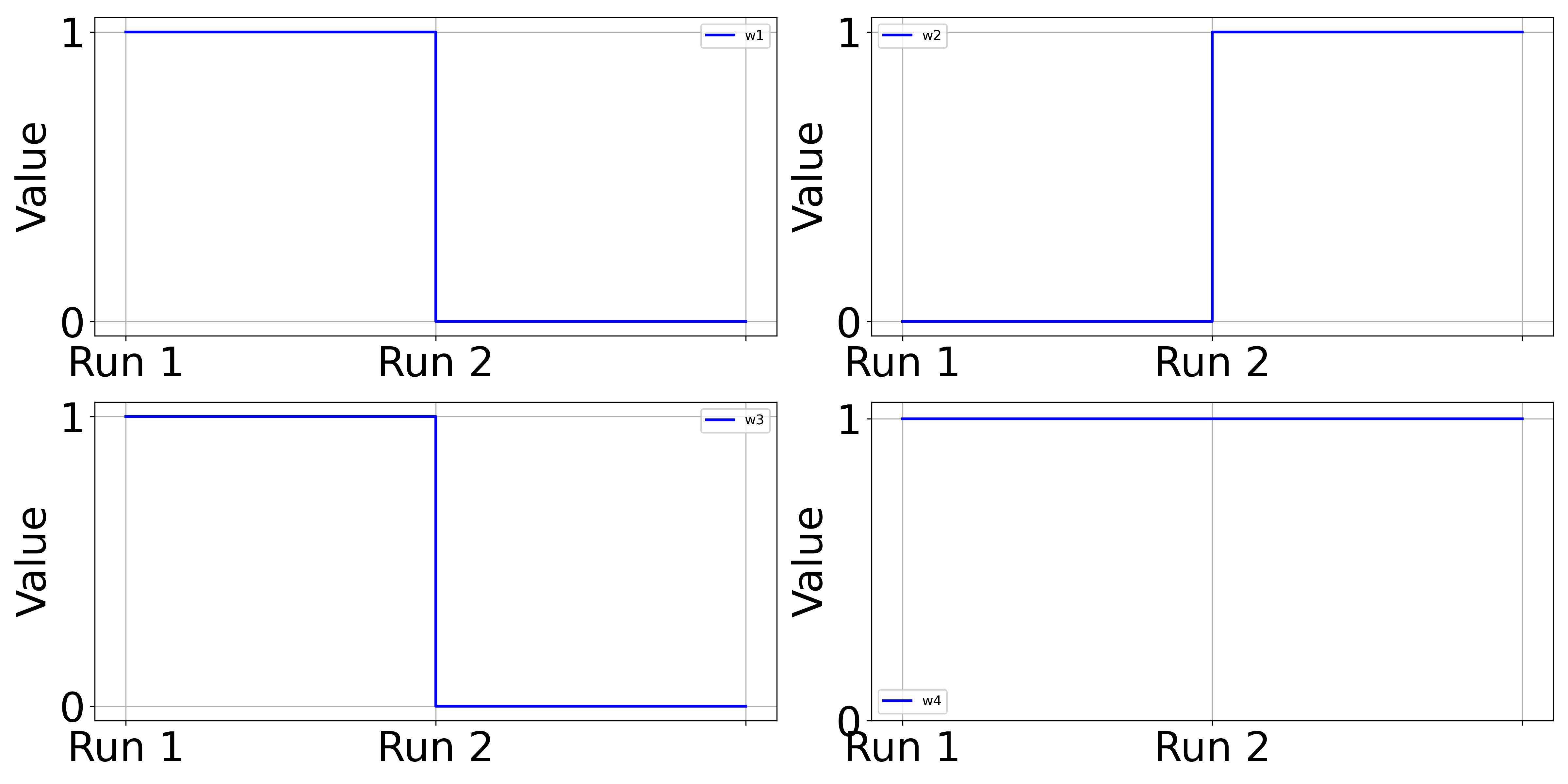}
  \caption{6.b}
  \label{fig:sfig2}
\end{subfigure}
\caption{This figure illustrates the comprehensive accuracy of our FL application in section 6.a, accounting for the primary run where all objective weights are equal, along with run 1 and run 2 as they relate to Figure 6.b.}
\label{fig:fig009}
\end{figure*}

3) In environments where static and mobile devices coexist, a treasure amount of data awaits the process of learning. However, many of these devices are frequently engaged or may lack the requisite capabilities to engage in FL. %This, in turn, results in delays and missed opportunities for ML applications. 
The strategic deployment of Docker containers through containerization technology becomes the catalyst for augmenting the available device pool, offering significant advantages to FL applications while elevating their accuracy as shown in Figure \ref{fig4}.
What sets our approach apart is the broad and diversified spectrum of potential clients it provides. %We commence with a small number of clients, progressively expanding it with each subsequent round from different area locations and environments.
By using our DRL solution, we swiftly attained our target accuracy in less than 10 rounds, marking a substantial advancement over the proficient but comparatively slower tabular RL approach. When confronted with a large number of devices, the tabular complexity of tabular RL significantly impacts memory and resources. In our exploration, we also evaluated a Genetic Algorithm. This algorithm often faces challenges in swiftly adapting to the rapid transformations inherent in dynamic environments. Notably, in the context of this implementation, heuristics tended to extend the convergence process, requiring more rounds to reach stability, particularly in new and rapidly changing area locations and environments. In contrast, our solution demonstrated exceptional performance by adeptly navigating and adapting to the dynamic nature of the environment. This stands in contrast to genetic algorithms, which may encounter challenges converging as efficiently due to the risk of falling into local optima, necessitating more rounds for the model to reach convergence. 
In contrast, the VanillaFL framework adheres to a fixed fraction of clients. The distinguishing feature of our architecture lies in its capacity to dynamically select the client set for each round, and efficiently adapt to changing conditions while using DRL agents. The expeditious convergence of the DRL algorithm underscores its effectiveness in addressing the challenges posed by swift and unpredictable changes in the environment, solidifying its position as a more adaptive and responsive solution. Reducing the number of rounds required for convergence holds several significant advantages. Firstly, it minimizes the communication overhead between the central server and individual clients, thereby reducing bandwidth consumption and latency. Furthermore, it also mitigates the risk of privacy breaches and reduces the computational burden on clients. Finally, it helps result in a faster model update and adaptation to changing data distributions or client populations.

4) Furthermore, our aim is to assess the effectiveness of our objective functions and their impact on the learning process. Figure \ref{fig:fig009} displays the performance of our FL application employing our DRL agent.
In the Main run, the agent underwent training with an equal weighting assigned to all of the objectives, ensuring a balanced and optimized deployment strategy. In contrast, during Run 1, the agent's training prioritized objectives $w_1$, $w_3$, and $w_4$, assigning them a weight of 1, while setting $w_2$ to zero. This setup resulted in the agent achieving a substantial accuracy level, reaching the desired threshold by round 14. A comparison with the Main Run reveals an additional 4 rounds needed to attain the same accuracy level. Notably, the delayed convergence can be attributed to the neglect of clients with substantial data volumes generated from their movements, suggesting that their inclusion could expedite convergence.

For Run 2, the agent's training focused on objectives $w_2$ and $w_4$, assigning them a weight of 1, while setting $w_1$ and $w_3$ to zero. This approach aimed to prioritize clients with high movement volumes and discard the objective that looks for clients with local accuracies similar to the global accuracy. The intention was to deploy these high-movement clients early in the process, potentially leading to higher global accuracy. However, Figure \ref{fig:fig009}.a indicates that this strategy prolonged the convergence time to around 18 rounds, with early-round accuracy noticeably lower compared to Run 1 and the Main run.
This discrepancy underscores the significance of including clients with comparable local accuracies to the global accuracy in the early rounds of learning. Such inclusion can significantly elevate the global accuracy in the initial stages of deployment. 
Additionally, when $w_1$ is set to zero, the number of deployed clients can exceed that of the Main Run, but with similar or lower overall accuracy. Consequently, more resources are expended without a corresponding increase in accuracy. This underscores the critical importance of selecting an optimal combination of quantity and quality of clients to achieve the desired accuracy with fewer learning rounds and minimized resource utilization. This is particularly important in On-Demand applications, where client devices may be repurposed for alternative deployments beyond the current task.

%We compared Our DQN algorithm with another DRL algorithms which is Proximal Policy Optimization (PPO) \cite{kl2}. PPO is known for its stability and ease of use. As shown in Figure \ref{fig8} we notice that in a highly changing environment, PPO was very sensitive to fast and dynamic changes and was not stable. The clipping mechanism in PPO controls the policy updates, providing some robustness to changes, but it may not respond as quickly as A2C. DQN tends to be more stable and less sensitive to immediate changes in the environment due to its experience replay buffer and target network mechanisms. 

\begin{figure}[]
	\centering
	\includegraphics[width=0.43\textwidth]{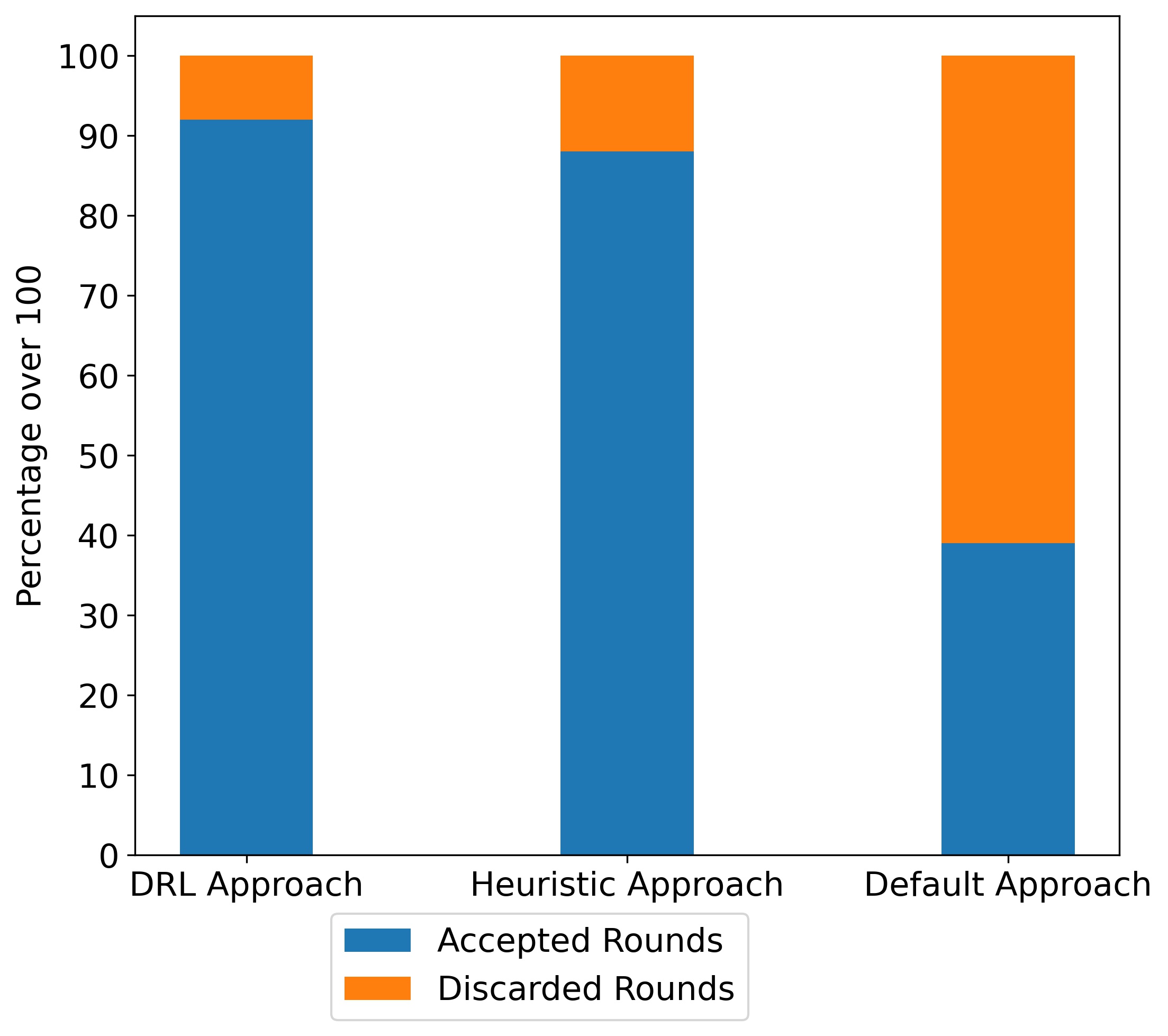}
	\caption{Percentage of Accepted and Discarded Rounds}
	\label{fig7}
\end{figure}

5) Furthermore, data availability and shifts significantly impact the total number of rounds required for model convergence. Client movements, influencing the data distribution the model learns from, directly affect access to available or new data, consequently impacting the overall convergence speed. This relationship is illustrated in Fig \ref{fig7}, where our DRL approach successfully utilized at least 92\% of the training rounds. In contrast, the heuristic approach may get stuck in local optima, and the random factors in selecting parent chromosomes introduce unpredictability in predicting data shifts, necessitating more rounds for convergence. The default approach, lacking effective selection, resulted in discarding over 60\% of rounds.% due to inefficiencies in responding to changes in the environment.
%\begin{figure}[]
%	\centering
%	\includegraphics[width=0.45\textwidth]{final2.png}
%	\caption{Cost evolution of DQN, PPO, and A2C over Time}
%	\label{fig8}
%\end{figure}

%\end{enumerate}

\section{Conclusion}

This paper focuses on dynamically deploying FL clients in specific regions through DRL and collaboration with volunteering devices, utilizing Kubeadm and Docker for swift response to real-time demands. 
Our approach takes into account the mobility of clients and performs optimally in environments with a high volume of available devices to be used. Our experiments illustrate the effectiveness of deploying On-Demand FL clients, substantially reducing the time required to achieve the desired accuracy compared to conventional methods. Simulations in real-world scenarios validate the feasibility of our approach, underscoring the advantages of incorporating more clients into the learning process. Our findings indicate improvements in accuracy during the initial stages, fewer discarded rounds, and increased availability of volunteering clients for other applications, resulting in reduced resource usage. Our approach underscores the importance of adapting to data shifts in FL settings by having new client deployments in areas where clients change locations. Moving forward, we aim to explore the integration of multiple models into our framework and incorporate reward-shaping techniques to further enhance the solution.

\ifCLASSOPTIONcaptionsoff
  \newpage
\fi

\bibliographystyle{IEEEtran}
\bibliography{references}

\begin{IEEEbiography}[{\includegraphics[width=1in,height=1.25in,clip]{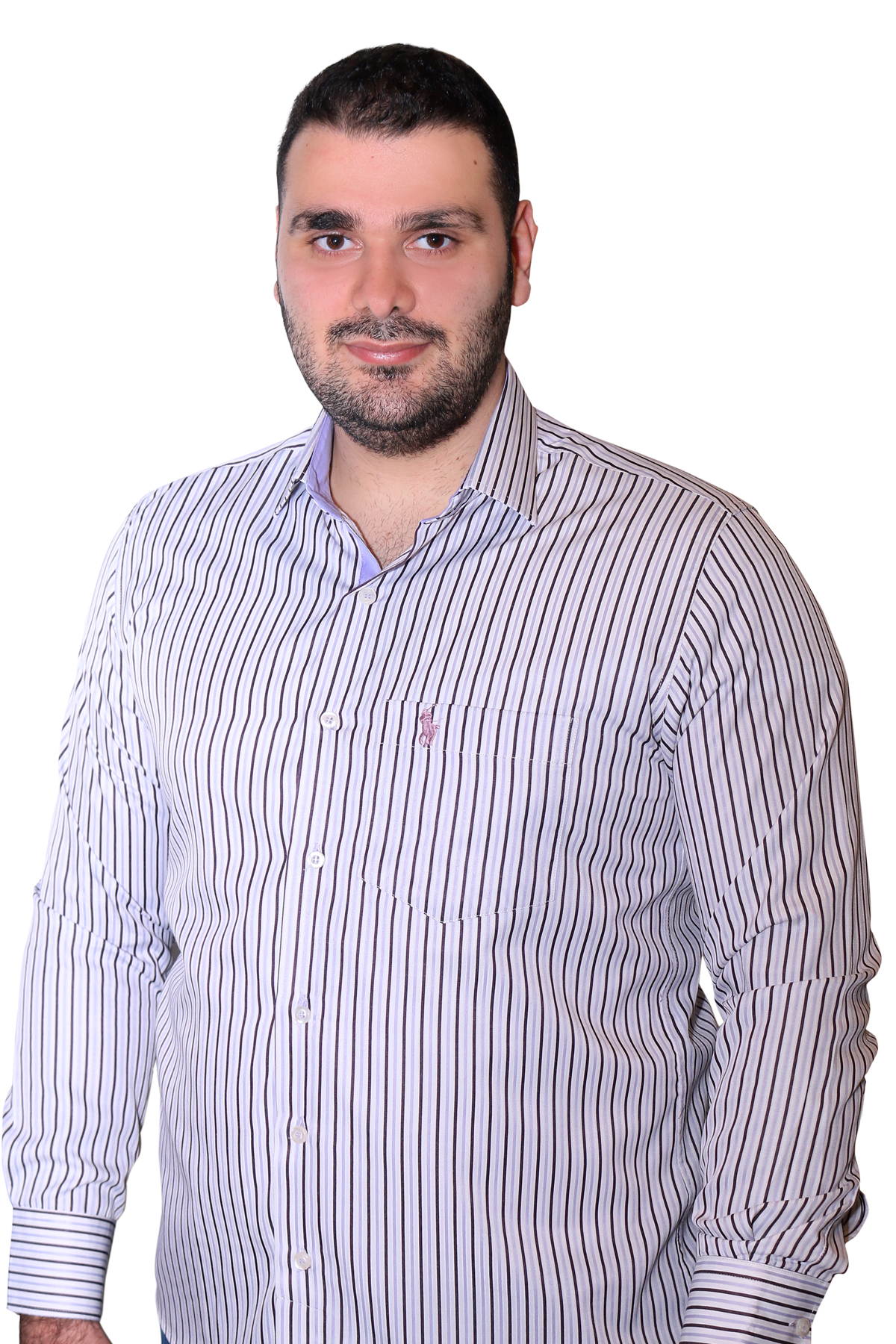}}]%
{\textbf{\textit{\textbf{Mario Chahoud}}}}
received his M.Sc. degree and B.S. degree in Computer Science from the Lebanese American University (LAU). He is currently a Research Fellow at the LAU Cyber Security Systems and Applied Artificial Intelligence Research Center and at Mohamad Bin Zayed University of Artificial Intelligence (MBZUAI), Abu Dhabi, United Arab Emirates. He was a Research and Teaching Assistant at the Lebanese American University. He is currently a Ph.D. student at Concordia university, Montreal, Canada. His current research interests include fog and cloud computing, Artificial intelligence, Machine learning, Federated Learning, and Cyber Security.
\end{IEEEbiography}
\vskip 0pt plus -1fil

\begin{IEEEbiography}[{\includegraphics[width=1in,height=1.25in,clip,keepaspectratio]{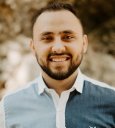}}]%
{\textbf{\textit{\textbf{Hani Sami}}}}
is currently a Ph.D. Candidate at Concordia University, Institute for information Systems Engineering (CIISE). He received his M.Sc. degree in Computer Science from the American University of Beirut  and completed his B.S. and worked as Research Assistant at the Lebanese American University. The topics of his research are Fog Computing, Vehicular Fog Computing, Reinforcement Learning, Reward Shaping, and Blockchain. He is a reviewer of several prestigious conferences and journals.
\end{IEEEbiography}
\vskip 0pt plus -1fil

\begin{IEEEbiography}[{\includegraphics[width=1in,height=1.25in,clip,keepaspectratio]{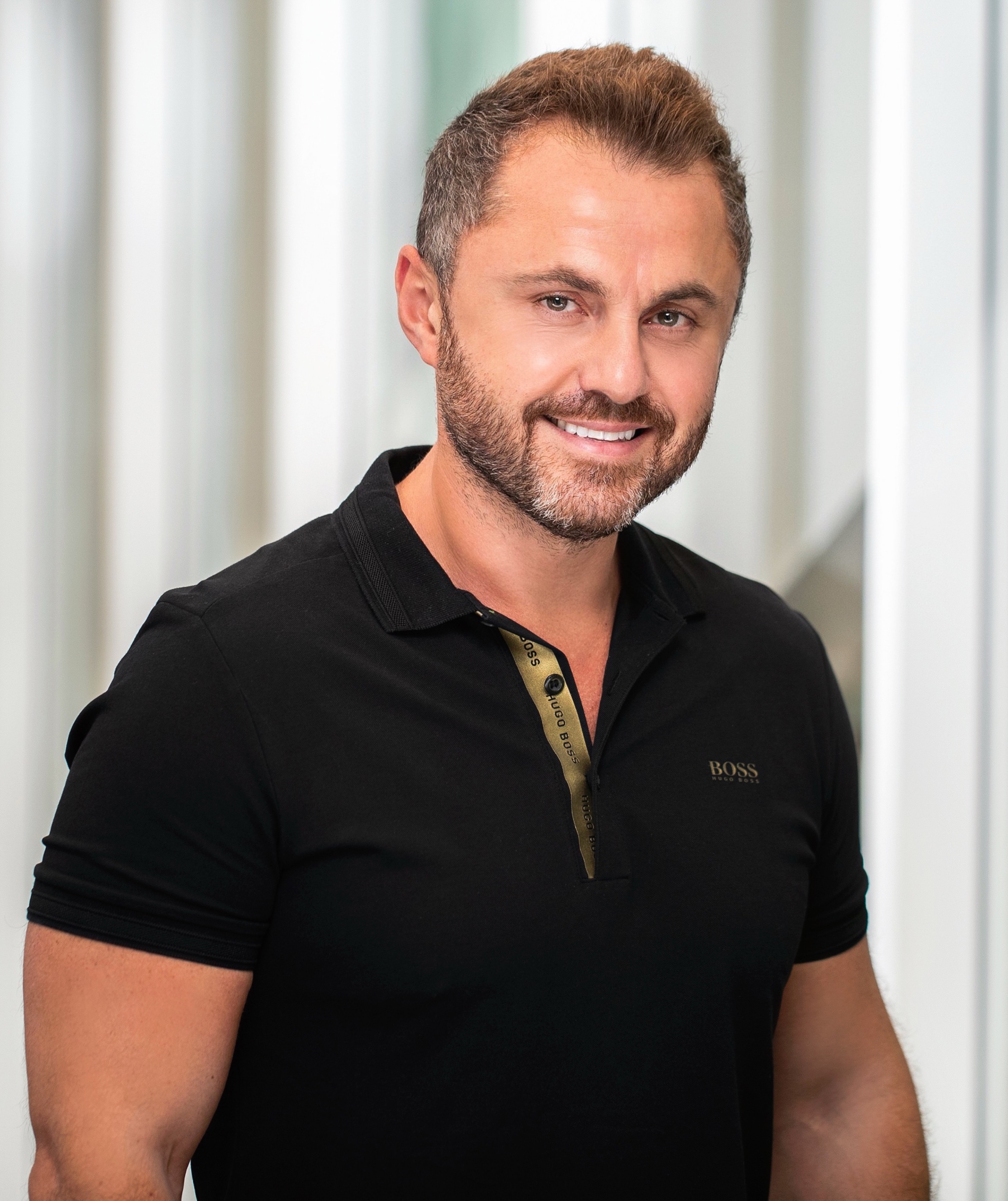}}]%
{\textbf{\textit{\textbf{Azzam Mourad}}}} is currently a Visiting Professor at Khalifa University, a Professor of Computer Science and Founding Director of the Artificial Intelligence and Cyber Systems Research Center at the Lebanese American University, and an Affiliate Professor at the Software Engineering and IT Department, Ecole de Technologie Superieure (ETS), Montreal, Canada. He was a Visiting Professor at New York University Abu Dhabi. His research interests include Cyber Security, Federated Machine Learning, Network and Service Optimization and Management targeting IoT and IoV, Cloud/Fog/Edge Computing, and Vehicular and Mobile Networks. He has served/serves as an associate editor for IEEE Transactions on Services Computing, IEEE Transactions on Network and Service Management, IEEE Network, IEEE Open Journal of the Communications Society, IET Quantum Communication, and IEEE Communications Letters, the General Chair of IWCMC2020-2022, the General Co-Chair of WiMob2016, and the Track Chair, a TPC member, and a reviewer for several prestigious journals and conferences. He is an IEEE senior member.
\end{IEEEbiography}

\vskip 0pt plus -1fil

\begin{IEEEbiography}[{\includegraphics[width=1in,height=1.25in,clip,keepaspectratio]{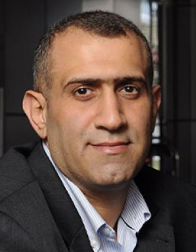}}]%
{\textbf{\textit{\textbf{Hadi Otrok}}}}
(senior member, IEEE) received his Ph.D. in ECE from Concordia University, Montreal, QC, Canada, in 2008.He holds a Full Professor position in the Department of Computer Science at Khalifa University, Abu Dhabi, UAE. He is also an Affiliate Associate Professor in the Concordia Institute for Information Systems Engineering at Concordia University, and an Affiliate Associate Professor in the Electrical Department at Ecole de Technologie Superieure (ETS), Montreal, Canada. His research interests include the domain of blockchain, reinforcement learning, federated learning, crowd sensing and sourcing, ad hoc networks, and cloud security. He co-chaired several committees at various IEEE conferences. He is also an Associate Editor at IEEE Transactions on Network and Service Management (TNSM), IEEE Transactions on Service Computing, and Ad-hoc networks (Elsevier). He also served in the editorial board of IEEE Networks and IEEE Communication Letters.
\end{IEEEbiography}
\vskip 0pt plus -1fil

\begin{IEEEbiography}[{\includegraphics[width=1in,height=1.25in,clip,keepaspectratio]{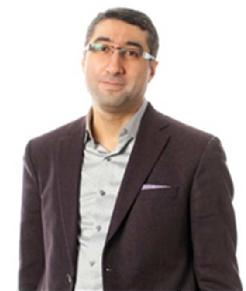}}]%
{\textbf{\textit{\textbf{Jamal Bentahar}}}}
received the Ph.D. degree in computer science and software engineering from Laval University, Canada, in 2005. He is a Professor with Concordia Institute for Information Systems Engineering, Concordia University, Canada and visiting professor at Khalifa University, UAE. From 2005 to 2006, he was a Postdoctoral Fellow with Laval University, and then NSERC Postdoctoral Fellow at Simon Fraser University, Canada. He was an NSERC Co-Chair for Discovery Grants for Computer Science (2016–2018). He is an associate editor of IEEE Transactions of Services Computing. His research interests include computational logics, model checking, reinforcement and deep learning, multi-agent systems, and services computing.
\end{IEEEbiography}

\vskip 0pt plus -1fil
\begin{IEEEbiography}[{\includegraphics[width=1in,height=1.25in,clip,keepaspectratio]{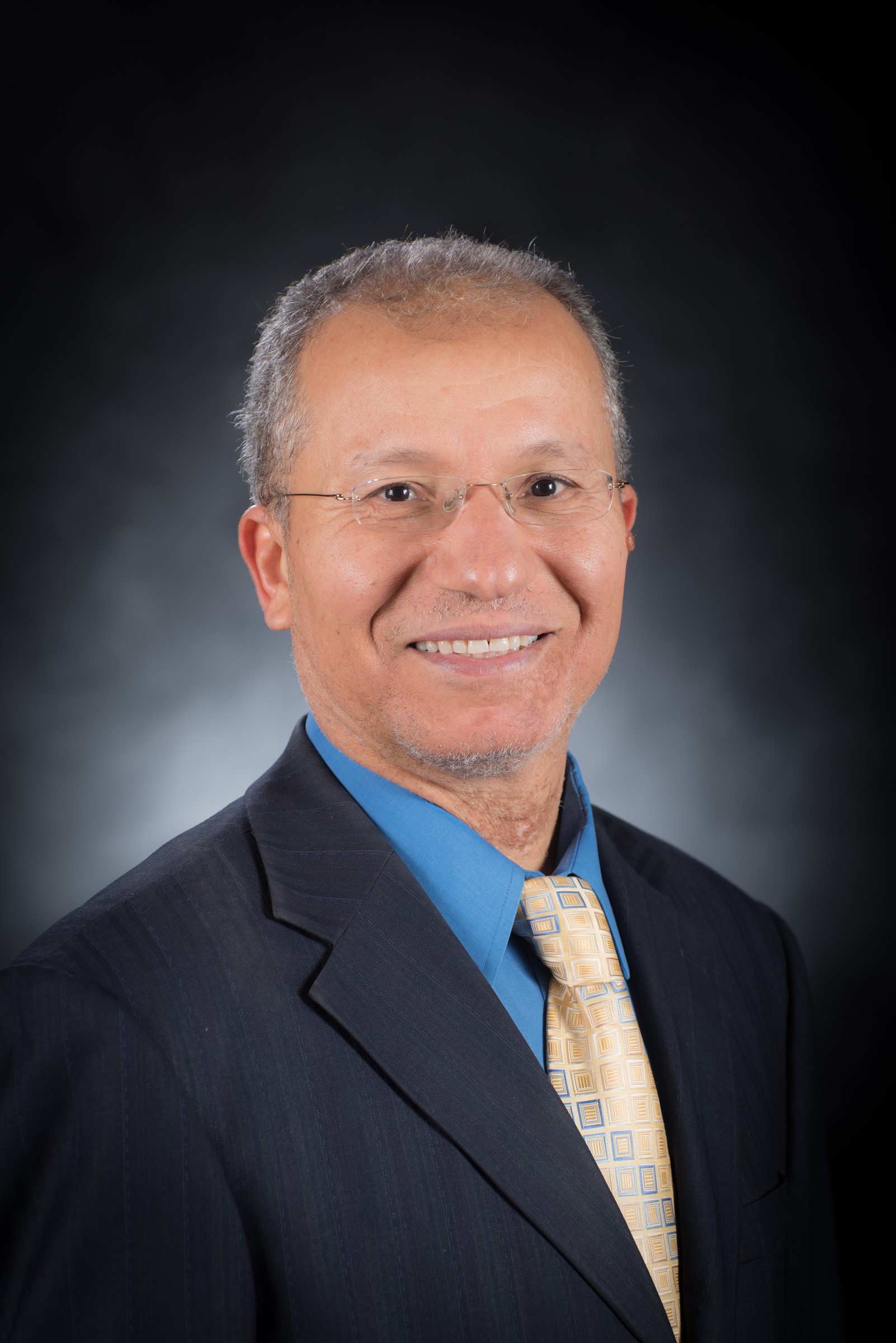}}]%
{\textbf{\textit{\textbf{Mohsen Guizani}}}}
Mohsen Guizani (Fellow, IEEE) received the BS (with distinction), MS and PhD degrees in Electrical and Computer engineering from Syracuse University, Syracuse, NY, USA in 1985, 1987 and 1990, respectively. He is currently a Professor of Machine Learning and the Associate Provost at Mohamed Bin Zayed University of Artificial Intelligence (MBZUAI), Abu Dhabi, UAE. Previously, he worked in different institutions in the USA. His research interests include applied machine learning and artificial intelligence, Internet of Things (IoT), intelligent autonomous systems, smart city, and cybersecurity. He was elevated to the IEEE Fellow in 2009 and was listed as a Clarivate Analytics Highly Cited Researcher in Computer Science in 2019, 2020 and 2021. Dr. Guizani has won several research awards including the “2015 IEEE Communications Society Best Survey Paper Award”, the Best ComSoc Journal Paper Award in 2021 as well five Best Paper Awards from ICC and Globecom Conferences. He is the author of ten books and more than 800 publications. He is also the recipient of the 2017 IEEE Communications Society Wireless Technical Committee (WTC) Recognition Award, the 2018 AdHoc Technical Committee Recognition Award, and the 2019 IEEE Communications and Information Security Technical Recognition (CISTC) Award. He served as the Editor-in-Chief of IEEE Network and is currently serving on the Editorial Boards of many IEEE Transactions and Magazines. He was the Chair of the IEEE Communications Society Wireless Technical Committee and the Chair of the TAOS Technical Committee. He served as the IEEE Computer Society Distinguished Speaker and is currently the IEEE ComSoc Distinguished Lecturer. 
\end{IEEEbiography}

\end{document}